\documentclass[11pt,a4paper]{article}
\usepackage[hyperref]{emnlp2018}
\usepackage{times}
\usepackage{CJK}
\usepackage{latexsym}
\usepackage{lipsum} 
\usepackage{amsmath}
\usepackage{graphicx}
\usepackage[english]{babel}
\usepackage{url}

\aclfinalcopy 
\def\aclpaperid{1120} 

\title{Unsupervised Neural Word Segmentation for Chinese\\ via Segmental Language Modeling}

\author{
  Zhiqing Sun \\
  Peking University \\
  {\tt 1500012783@pku.edu.cn} \\\And
  Zhi-Hong Deng \\
  Peking University \\
  {\tt zhdeng@pku.edu.cn}\\
}
\date{}

\begin{document}
\maketitle
\begin{abstract}
Previous traditional approaches to unsupervised Chinese word segmentation (CWS) can be roughly classified into discriminative and generative models. The former uses the carefully designed goodness measures for candidate segmentation, while the latter focuses on finding the optimal segmentation of the highest generative probability. However, while there exists a trivial way to extend the discriminative models into neural version by using neural language models, those of generative ones are non-trivial. In this paper, we propose the segmental language models (SLMs) for CWS. Our approach explicitly focuses on the segmental nature of Chinese, as well as preserves several properties of language models. In SLMs, a context encoder encodes the previous context and a segment decoder generates each segment incrementally. As far as we know, we are the first to propose a neural model for unsupervised CWS and achieve competitive performance to the state-of-the-art statistical models on four different datasets from SIGHAN 2005 bakeoff.
\end{abstract}

\begin{figure}[t!]
\center
\includegraphics[width=6 cm]{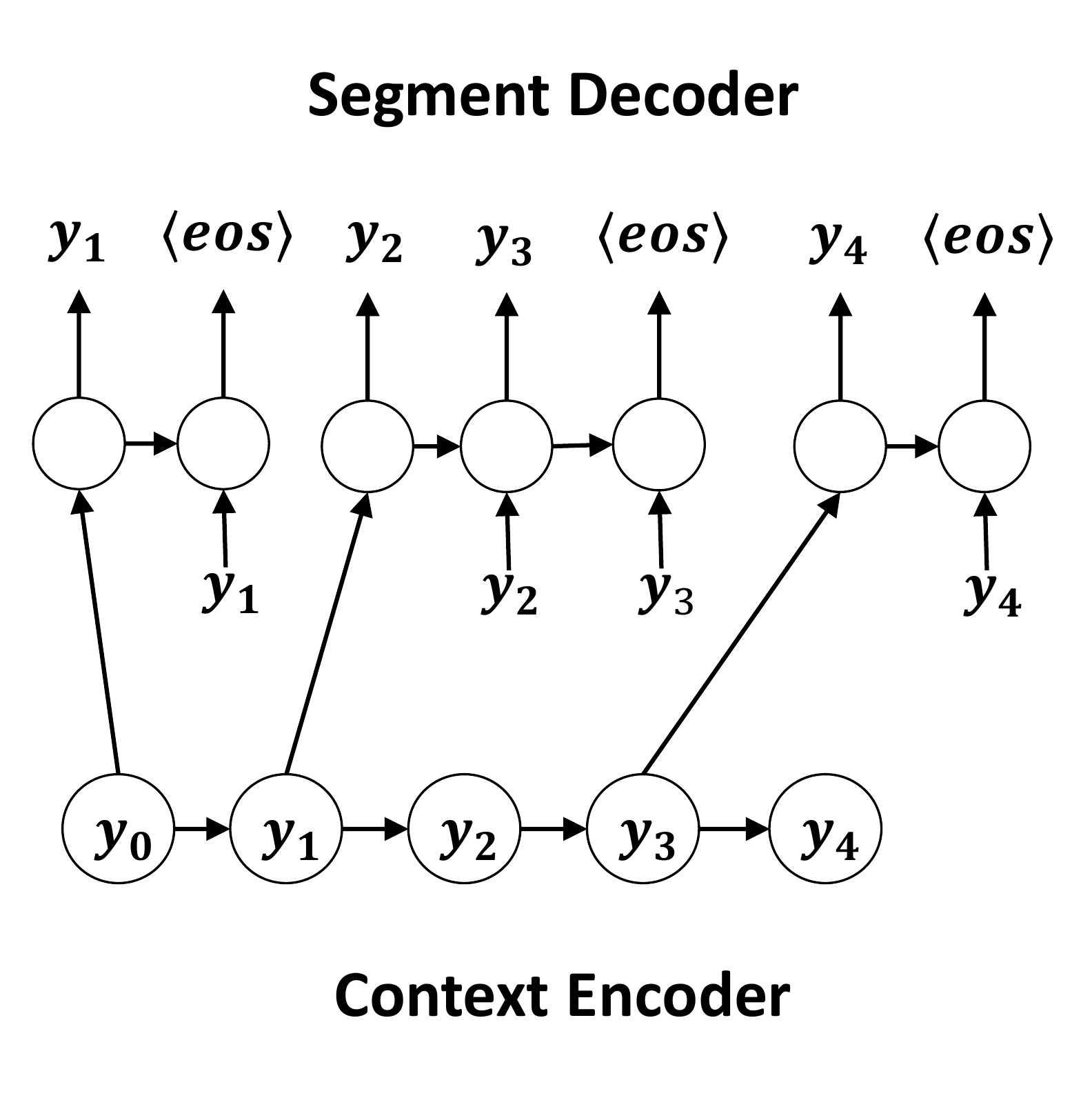}
\caption{A Segmental Language Model (SLM) works on $\mathbf{y} = \mathbf{y}_1\mathbf{y}_2\mathbf{y}_3\mathbf{y}_4$ with the candidate segmentation $\mathbf{y}_1$, $\mathbf{y}_{2:3}$ and $\mathbf{y}_4$, where $\mathbf{y}_0$ is an additional start symbol which is kept same for all sentences.}
\label{fig:SLM}
\end{figure}

\section{Introduction}

Unlike English and many other languages, Chinese sentences have no explicit word boundaries. Therefore, Chinese Word Segmentation (CWS) is a crucial step for many Chinese Natural Language Processing (NLP) tasks such as syntactic parsing, information retrieval and word representation learning \citep{grave2018learning}.

Recently, neural approaches for supervised CWS are attracting huge interest. A great quantities of neural models, e.g., tensor neural network  \citep{pei-ge-chang:2014:P14-1}, recursive neural network \citep{chen-EtAl:2015:ACL-IJCNLP5}, long-short-term-memory (RNN-LSTM) \citep{chen-EtAl:2015:EMNLP2} and convolutional neural network (CNN) \citep{Wang:2017}, have been proposed and given competitive results to the best statistical models \citep{sun:2010:POSTERS}. However, the neural approaches for unsupervised CWS have not been investigated.

Previous unsupervised approaches to CWS can be roughly classified into discriminative and generative models. The former uses carefully designed goodness measures for candidate segmentation, while the latter focuses on designing statistical models for Chinese and finds the optimal segmentation of the highest generative probability.

Popular goodness measures for discriminative models include Mutual Information (MI) \citep{chang2003unsupervised}, normalized Variation of Branching Entropy (nVBE) \citep{magistry2012unsupervized} and Minimum Description Length (MDL) \citep{magistry2013can}. There is a trivial way to extend these statistical discriminative approaches, because we can simply replace the n-gram language models in these approaches by neural language models \citep{bengio2003neural}. There may exists other more sophisticated neural discriminative approaches, but it is not the focus of this paper.

For generative approaches, typical statistical models includes Hidden Markov Model (HMM) \citep{chen-chang-pei:2014:EMNLP2014}, Hierarchical Dirichlet Process (HDP) \citep{goldwater2009bayesian} and Nested Pitman-Yor Process (NPY) \citep{mochihashi2009bayesian}. However, none of them can be easily extended into a neural model. Therefore, neural generative models for word segmentation are remaining to be investigated.

In this paper, we proposed the Segmental Language Models (SLMs), a neural generative model that explicitly focuses on the segmental nature of Chinese: SLMs can directly generate segmented sentences and give the corresponding generative probability. We evaluate our methods on four different benchmark datasets from SIGHAN 2005 bakeoff \citep{emerson2005second}, namely PKU, MSR, AS and CityU. To our knowledge, we are the first to propose a neural model for unsupervised Chinese word segmentation and achieve competitive performance to the state-of-the-art statistical models on four different datasets.\footnote{Our implementation can be found at \url{https://github.com/Edward-Sun/SLM}}
    
\section{Segmental Language Models}

In this section, we present our segmental language models (SLMs). Notice that in Chinese NLP, characters are the atom elements. Thus in the context of CWS, we use ``character'' instead of ``word'' for language modeling.

\subsection{Language Models}

The goal of language modeling is to learn the joint probability function of sequences of characters in a language. However, This is intrinsically difficult because of the curse of dimensionality. Traditional approaches obtain generalization based on n-grams, while neural approaches introduce a distributed representation for characters to fight the curse of dimensionality.

A neural Language Model (LM) can give the conditional probability of the next character given the previous ones, and is usually implemented by a Recurrent Neural Network (RNN):

\begin{align}
	&\mathbf{h}_t = f(\mathbf{y}_{t-1}, \mathbf{h}_{t-1})\\
	& p(\mathbf{y}_t | \mathbf{y}_{1:t-1}) = g(\mathbf{h}_t, \mathbf{y}_t)
\end{align}

where $\mathbf{y}_t$ is the distributed representation for the $t^{th}$ character and $\mathbf{h}_t$ represents the information of the previous characters.

\subsection{Segmental Language Models}

Similar to neural language modeling, the goal of segmental language modeling is to learn the joint probability function of the segmented sequences of characters. Thus, for each segment, we have:

\begin{equation}
	\hat{p}(\mathbf{y}_t^{(i)} | \mathbf{y}_{1:t-1}^{(i)}, \mathbf{y}^{(1:i-1)}) = g(\mathbf{h}_t^{(i)}, \mathbf{y}_t^{(i)})
\end{equation}

where $\mathbf{y}_t^{(i)}$ is the distributed representation for the $t^{th}$ character in the $i^{th}$ segment and $\mathbf{y}^{(1:i-1)}$ is the previous segments. And the concatenation of all segments $\mathbf{y}_{1:T_i}^{(i)}$ is exactly the whole sentence $\mathbf{y}_{1:T}$, where $T_i$ is the length of the $i^{th}$ segment $\mathbf{y}^{(i)}$, $T$ is the length of the sentence $\mathbf{y}$.

Moreover, we introduce a context encoder RNN to process the character sequence $\mathbf{y}^{(1:i-1)}$ in order to make $\mathbf{y}_t^{(i)}$ conditional on $\mathbf{y}^{(1:i-1)}$. Specifically, we initialize $\mathbf{h}_0^{(i)}$ with the context encoder's output of $\mathbf{y}^{(1:i-1)}$.

Notice that although we have an encoder and the segment decoder $g$, SLM is not an encoder-decoder model. Because the content that the decoder generates is not the same as what the encoder provides.

Figure \ref{fig:SLM} illustrates how SLMs work with a candidate segmentation.

\subsection{Properties of SLMs}

However, in unsupervised scheme, the given sentences are not segmented. Therefore, the probability for SLMs to generate a given sentence is the joint probability of all possible segmentation:

\begin{align}
	p(\mathbf{y}_{1:T}) &= \sum_{T_1, T_2, \dots} \prod_{i} \hat{p}(\mathbf{y}_{1:T_i}^{(i)}) \nonumber\\ 
    &= \sum_{T_1, T_2, \dots} \prod_{i} \prod_{t = 1}^{T_i + 1} \hat{p}(\mathbf{y}_t^{(i)} | \mathbf{y}_{0:t-1}^{(i)})
\end{align}

where $\mathbf{y}_{T_i+1}^{(i)} = \langle eos \rangle$ is the \textbf{end of segment} symbol at the end of each segment, and $\mathbf{y}_{0}^{(i)}$ is the context representation of $\mathbf{y}^{(1:i-1)}$.

Moreover, for sentence generation, SLMs are able to generate arbitrary sentences by generating segments one by one and stopping when generating \textbf{end of sentence} symbol $\langle EOS \rangle$. In addition, the time complexity is linear to the length of the generated sentence, as we can keep the hidden state of the context encoder RNN and update it when generating new words.

Last but not least, it is easy to verify that SLMs preserve the probabilistic property of language models:
\begin{equation}
\sum_i P(\mathbf{s}_i) = 1
\end{equation}
where $\mathbf{s}_i$ enumerates all possible sentences.

In summary, the segmental language models can perfectly substitute vanilla language models.

\subsection{Training and Decoding}

Similar to language model, the training is achieved by maximizing the training corpus log-likelihood:
\begin{equation}
	L =  - \log p(\mathbf{y}_{1:T})
\end{equation}

Luckily, we can compute the loss objective function in linear time complexity using dynamic programming, given the initial condition that $p(\mathbf{y}_{1:0}) = 1$:

\begin{equation}
	p(\mathbf{y}_{1:n}) = \sum_{k = 1}^{K} p(\mathbf{y}_{1:n-k}) \hat{p}(\mathbf{y}_{n-k+1:n})
\end{equation}

where $p(\cdot)$ is the joint probability of all possible segmentation, $\hat{p}(\cdot)$ is the probability of one segment and $K$ is the maximal length of the segments.

We can also find the segmentation with maximal probability (namely, decoding) in linear time using dynamic programming in the similarly way with $\bar{p}(\mathbf{y}_{1:0}) = 1$:

\begin{align}
	&\bar{p}(\mathbf{y}_{1:n}) = \max_{k = 1}^{K} \bar{p}(\mathbf{y}_{1:n-k}) \hat{p}(\mathbf{y}_{n-k+1:n})\\
    &\delta(\mathbf{y}_{1:n}) = \arg \max_{k = 1}^{K} \bar{p}(\mathbf{y}_{1:n-k}) \hat{p}(\mathbf{y}_{n-k+1:n})
\end{align}

where $\bar{p}$ is the probability of the best segmentation and $\delta$ is used to trace back the decoding.

\section{Experiments}

\subsection{Experimental Settings and Detail}

We evaluate our models on SIGHAN 2005 bakeoff \citep{emerson2005second} datasets and replace all the punctuation marks with $\langle punc \rangle$, English characters with $\langle eng \rangle$ and Arabic numbers with $\langle num \rangle$ for all text and only consider segment the text between punctuations. Following \citet{chen-chang-pei:2014:EMNLP2014} , we use both training data and test data for training and only test data are used for evaluation. In order to make a fair comparison with the previous works, we do not consider using other larger raw corpus. 

We apply word2vec \citep{mikolov2013distributed} on Chinese Gigaword corpus (LDC2011T13) to get pretrained embedding of characters.

A 2-layer LSTM \citep{hochreiter1997long} is used as the segment decoder and a 1-layer LSTM is used as the context encoder.

We use stochastic gradient decent with a mini-batch size of 256 and a learning rate of 16.0 to optimize the model parameters in the first 400 steps, then we use Adam \citep{kingma2014adam} with a learning rate of 0.005 to further optimize the models. Model parameters are initialized by normal distributions as \citet{glorot2010understanding} suggested. We use a gradient clip $= 0.1$ and apply a dropout with dropout rate $= 0.1$ to the character embedding and RNNs to prevent over-fit.

The standard word precision, recall and F1 measures \citep{emerson2005second} are used to evaluate segmentation performance.

\begin{table}[t]
\centering
 \begin{tabular}{|c|c|c|c|c|c|c|} 
 \hline
\textbf{F1 score} & \textbf{PKU} & \textbf{MSR} & \textbf{AS} &\textbf{CityU}\\
 \hline
 HDP & 68.7 & 69.9 & - & -\\
 \hline
 HDP + HMM & 75.3 & 76.3 & - & -\\
 \hline
 ESA & 77.8 & 80.1 & 78.5 & 76.0\\
 \hline
 NPY-3 & - & 80.7 & - & 81.7\\
 \hline
 NPY-2 & - & 80.2 & - & \textbf{82.4}\\
 \hline
 nVBE & 80.0 & \textbf{81.3} & 76.6 & 76.7\\
 \hline
 Joint & 81.1 & 81.7 & - & -\\
 \hline
 \hline
 SLM-2 & \textbf{80.2} & 78.5 & 79.4 & 78.2\\
 \hline
 SLM-3 & 79.8 & 79.4 & \textbf{80.3} & 80.5\\
 \hline
 SLM-4 & 79.2 & 79.0 & 79.8 & 79.7\\
 \hline
 \end{tabular}
\caption{Main results on SIGHAN 2005 bakeoff datasets with previous state-of-the-art models \citep{chen-chang-pei:2014:EMNLP2014, wang2011new, mochihashi2009bayesian, magistry2012unsupervized}}
\label{tab:results}
\end{table}

\subsection{Results and Analysis}

\begin{table}[t]
\centering
 \begin{tabular}{|c|c|c|c|c|c|c|} 
 \hline
\textbf{F1 score} & \textbf{PKU} & \textbf{MSR} & \textbf{AS} &\textbf{CityU}\\
\hline
 SLM-4 & 79.2 & 79.0 & 79.8 & 79.7\\
 \hline
 SLM-4* & 81.9 & 83.0 & 81.0 & 81.4\\
 \hline
  SLM-4\dag &  \textbf{87.5} & 84.3 & \textbf{84.2} & \textbf{86.0}\\
 \hline
  SLM-4\dag* &  87.3 & \textbf{84.8} &  83.9& 85.8\\
 \hline
 \end{tabular}
\caption{Results of SLM-4 incorporating ad hoc guidelines, where \dag  \ represents using additional 1024 segmented setences for training data and * represents using a rule-based post-processing}
\label{tab:improve}
\end{table}

Our final results are shown in Table \ref{tab:results}, which lists the results of several previous state-of-the-art methods\footnote{\citet{magistry2012unsupervized} evaluated their nVBE on the training data, and the joint model of \citet{chen-chang-pei:2014:EMNLP2014} combine HDP+HMM and is initialized with nVBE, so in principle these results can not be compared directly.}, where we mark the best results in \textbf{boldface}. We test the proposed SLMs with different maximal segment length $K = 2,3,4$ and use ``SLM-$K$'' to denote the corresponding model. We do not try $K > 4$ because there are rare words that consist more than 4 characters.

As can be seen, it is hard to predict what choice of $K$ will give the best performance. This is because the exact definition of what a word remains hard to reach and different datasets follow different guidelines. \citet{zhao2008empirical} use cross-training of a supervised segmentation system in order to have an estimation of the consistency between different segmentation guidelines and the average consistency is found to be as low as 85 (f-score). Therefore, this can be regarded as a top line for unsupervised CWS.

Table \ref{tab:results} shows that SLMs outperform previous best discriminative and generative models on PKU and AS datasets. This might be due to that the segmentation guideline of our models are closer to these two datasets.

\begin{CJK}{UTF8}{gbsn}
Moreover, in the experiments, we observe that Chinese particles often attach other words, for example, ``的'' following adjectives and ``了'' following verbs. It is hard for our generative models to split them apart. Therefore, we propose a rule-based post-processing module to deal with this problem, where we explicitly split the attached particles from other words.\footnote{The rules we use are listed in the appendix at \url{https://github.com/Edward-Sun/SLM}.} The post-processing is applied on the results of ``SLM-4''. In addition, we also evaluate ``SLM-4'' using the first 1024 sentences of the segmented training datasets (about 5.4\% of PKU, 1.2\% of MSR, 0.1\% of AS and 1.9\% of CityU) for training, in order to teach ``SLM-4'' the corresponding ad hoc segmentation guidelines. Table \ref{tab:improve} shows the results. 
\end{CJK}

We can find from the table that only 1024 guideline sentences can improve the performance of ``SLM-4'' significantly. While rule-based post-processing is very effective, ``SLM-4\dag'' can outperform ``SLM-4*'' on all the four datasets. Moreover, performance drops when applying the rule-based post-processing to ``SLM-4\dag'' on three datasets. These indicate that SLMs can learn the empirical rules for word segmentation given only a small amount of training data. And these guideline data can improve the performance of SLMs naturally, superior to using explicit rules.

\subsection{The Effect of the Maximal Segment Length}

The maximal segment length $K$ represents the prior knowledge we have for Chinese word segmentation. For example $K = 3$ represents that there are only unigrams, bigrams and trigrams in the text. While there do exist words that contain more than four characters, most of the Chinese words are unigram or bigram. Therefore, $K$ denotes a trade-off between the accuracy of short words and long words.

Specifically, we investigate two major segmentation problems that might affect the accuracy of word segmentation performance, namely, insertion errors and deletion errors. An insertion error insert a segment in a word, which split a correct word. And an deletion error delete the segment between two words, which results in a composition error \citep{li1998chinese}. Table \ref{tab:error} shows the statistics of different errors on PKU of our model with different $K$. We can observe that insertion error rate decrease with the increase of $K$, while the deletion error rate increase with the increase of $K$.

\begin{CJK}{UTF8}{gbsn}
We also provide some examples in Table \ref{tab:example}, which are taken from the results of our models. It clearly illustrates that different $K$ could result in different errors. For example, there is an insertion error on ``反过来'' by SML-2, and a deletion error on ``促进'' and ``了'' by SLM-4.
\end{CJK}

\begin{table}[t]
\centering
 \begin{tabular}{|c|c|c|c|c|c|c|} 
 \hline
 \textbf{Error} & \textbf{SLM-2} & \textbf{SLM-3} & \textbf{SLM-4}\\
 \hline
 Insertion & 7866 & 4803 & 3519\\
\hline
 Deletion & 3855 & 7518 & 8851\\
 \hline
 \end{tabular}
\caption{Statistics of insertion errors and deletion errors that SLM-$K$ produces on PKU dataset}
\label{tab:error}
\end{table}

\begin{table*}[t]
\centering
 \begin{CJK}{UTF8}{gbsn}
 \begin{tabular}{|c|c|c|c|c|c|c|}
 \hline
 \textbf{Model} & \textbf{Example}\\
 \hline
 SLM-2 & 而 这些 制度 的 完善 反 过来 又 促进 了 检察 人员 执法 水平 的 进一 步 提高 \\
\hline
 SLM-3 & 而 这些 制度 的 完善 反过来 又 促进了 检察 人员 执法 水平 的 进一步 提高 \\
 \hline
 SLM-4 & 而 这些 制度 的 完善 反过 来 又 促进了 检察 人员 执法 水平 的进一步 提高 \\
 \hline
 Gold & 而 这些 制度 的 完善 反过来 又 促进 了 检察 人员 执法 水平 的 进一步 提高\\
 \hline
 \end{tabular}
 \end{CJK}
\caption{Examples of segmentation with different maximal segment length $K$}
\label{tab:example}
\end{table*}

\section{Related Work}

\paragraph{Generative Models for CWS}
\citet{goldwater2009bayesian} are the first to proposed a generative model for unsupervised word segmentation. They built a nonparametric Bayesian bigram language model based on HDP \citep{teh2005sharing}. \citet{mochihashi2009bayesian} proposed a Bayesian hierarchical language model using Pitman-Yor (PY) process, which can generate sentences hierarchically. \citet{chen-chang-pei:2014:EMNLP2014} proposed a Bayesian HMM model for unsupervised CWS inspired by the character-based scheme in supervised CWS task, where the hidden state of charaters are set to $\{ \mathbf{S}\text{ingle}, \mathbf{B}\text{egin}, \mathbf{E}\text{nd}, \mathbf{M}\text{iddle} \}$ to represents their corresponding positions in the words. The segmental language model is not a neural extension of the above statistical models, as we model the segments directly.

\paragraph{Segmental Sequence Models}
Sequence modeling via segmentations has been well investigated by \citet{wang2017sequence}, where they proposed the Sleep-AWake Network (SWAN) for speech recognition. SWAN is similar to SLM. However, SLMs do not have sleep-awake states. And SLMs predict the following segment given the previous context while SWAN tries to recover the information in the encoded state. Therefore, the key difference is that SLMs are unsupervised language models while SWANs are supervised seq2seq models. Thereafter, \citet{huang2017towards} successfully apply SWAN in their phrase-based machine translation. Another related work in machine translation is the online segment to segment neural transduction \citep{yu2016online}, where the model is able to capture unbounded dependencies in both the input and output sequences. \citet{kong2017neural} also proposed a Segmental Recurrent Neural Network (SRNN) with CTC to solve segmental labeling problems.

\section{Conclusion}

In this paper, we proposed a neural generative model for fully unsupervised Chinese word segmentation (CWS). To the best of knowledge, this is the first neural model for CWS. Our segmental language model is an intuitive generalization of vanilla neural language models that directly modeling the segmental nature of Chinese. Experimental results show that our models achieve competitive performance to the previous state-of-the-art statistical models on four datasets from SIGHAN 2005. We also show the improvement of incorporating ad hoc guidelines into our segmental language models. Our future work may include the following two directions.
\begin{itemize}
\item In this work, we only consider the sequential segmental language modeling.  In the future, we are interested in build a hierarchical neural language model like the Pitman-Yor process.
\item Like vanilla language models, the segmental language models can also provide useful information for semi-supervised learning tasks.  It would also be interesting to explore our models in the semi-supervised schemes.
\end{itemize}

\section*{Acknowledgements}
This work is supported by the National Training Program of Innovation for Undergraduates (URTP2017PKU001). We would also like to thank the anonymous reviewers for their helpful comments.

\bibliography{CWS}

\begin{thebibliography}{26}
\expandafter\ifx\csname natexlab\endcsname\relax\def\natexlab#1{#1}\fi

\bibitem[{Bengio et~al.(2003)Bengio, Ducharme, Vincent, and
  Jauvin}]{bengio2003neural}
Yoshua Bengio, R{\'e}jean Ducharme, Pascal Vincent, and Christian Jauvin. 2003.
\newblock A neural probabilistic language model.
\newblock \emph{Journal of machine learning research}, 3(Feb):1137--1155.

\bibitem[{Chang and Lin(2003)}]{chang2003unsupervised}
Jason~S Chang and Tracy Lin. 2003.
\newblock Unsupervised word segmentation without dictionary.
\newblock \emph{ROCLING 2003 Poster Papers}, pages 355--359.

\bibitem[{Chen et~al.(2014)Chen, Chang, and
  Pei}]{chen-chang-pei:2014:EMNLP2014}
Miaohong Chen, Baobao Chang, and Wenzhe Pei. 2014.
\newblock A joint model for unsupervised chinese word segmentation.
\newblock In \emph{Proceedings of the 2014 Conference on Empirical Methods in
  Natural Language Processing (EMNLP)}, pages 854--863, Doha, Qatar.
  Association for Computational Linguistics.

\bibitem[{Chen et~al.(2015{\natexlab{a}})Chen, Qiu, Zhu, and
  Huang}]{chen-EtAl:2015:ACL-IJCNLP5}
Xinchi Chen, Xipeng Qiu, Chenxi Zhu, and Xuanjing Huang. 2015{\natexlab{a}}.
\newblock Gated recursive neural network for chinese word segmentation.
\newblock In \emph{Proceedings of the 53rd Annual Meeting of the Association
  for Computational Linguistics and the 7th International Joint Conference on
  Natural Language Processing (Volume 1: Long Papers)}, pages 1744--1753,
  Beijing, China. Association for Computational Linguistics.

\bibitem[{Chen et~al.(2015{\natexlab{b}})Chen, Qiu, Zhu, Liu, and
  Huang}]{chen-EtAl:2015:EMNLP2}
Xinchi Chen, Xipeng Qiu, Chenxi Zhu, Pengfei Liu, and Xuanjing Huang.
  2015{\natexlab{b}}.
\newblock Long short-term memory neural networks for chinese word segmentation.
\newblock In \emph{Proceedings of the 2015 Conference on Empirical Methods in
  Natural Language Processing}, pages 1197--1206, Lisbon, Portugal. Association
  for Computational Linguistics.

\bibitem[{Emerson(2005)}]{emerson2005second}
Thomas Emerson. 2005.
\newblock The second international chinese word segmentation bakeoff.
\newblock In \emph{Proceedings of the fourth SIGHAN workshop on Chinese
  language Processing}, volume 133, pages 123--133.

\bibitem[{Glorot and Bengio(2010)}]{glorot2010understanding}
Xavier Glorot and Yoshua Bengio. 2010.
\newblock Understanding the difficulty of training deep feedforward neural
  networks.
\newblock In \emph{Proceedings of the Thirteenth International Conference on
  Artificial Intelligence and Statistics}, pages 249--256.

\bibitem[{Goldwater et~al.(2009)Goldwater, Griffiths, and
  Johnson}]{goldwater2009bayesian}
Sharon Goldwater, Thomas~L Griffiths, and Mark Johnson. 2009.
\newblock A bayesian framework for word segmentation: Exploring the effects of
  context.
\newblock \emph{Cognition}, 112(1):21--54.

\bibitem[{Grave et~al.(2018)Grave, Bojanowski, Gupta, Joulin, and
  Mikolov}]{grave2018learning}
Edouard Grave, Piotr Bojanowski, Prakhar Gupta, Armand Joulin, and Tomas
  Mikolov. 2018.
\newblock Learning word vectors for 157 languages.
\newblock In \emph{Proceedings of the International Conference on Language
  Resources and Evaluation (LREC 2018)}.

\bibitem[{Hochreiter and Schmidhuber(1997)}]{hochreiter1997long}
Sepp Hochreiter and J{\"u}rgen Schmidhuber. 1997.
\newblock Long short-term memory.
\newblock \emph{Neural computation}, 9(8):1735--1780.

\bibitem[{Huang et~al.(2017)Huang, Wang, Huang, Zhou, and
  Deng}]{huang2017towards}
Po-Sen Huang, Chong Wang, Sitao Huang, Dengyong Zhou, and Li~Deng. 2017.
\newblock Computer science > computation and language towards neural
  phrase-based machine translation.
\newblock \emph{arxiv.org/abs/1706.05565}.

\bibitem[{Kingma and Ba(2014)}]{kingma2014adam}
Diederik Kingma and Jimmy Ba. 2014.
\newblock Adam: A method for stochastic optimization.
\newblock \emph{arXiv preprint arXiv:1412.6980}.

\bibitem[{Kong(2017)}]{kong2017neural}
Lingpeng Kong. 2017.
\newblock \emph{Neural Representation Learning in Linguistic Structured
  Prediction}.
\newblock Ph.D. thesis, Google Research.

\bibitem[{Li and Yuan(1998)}]{li1998chinese}
Haizhou Li and Baosheng Yuan. 1998.
\newblock Chinese word segmentation.
\newblock In \emph{Proceedings of the 12th Pacific Asia Conference on Language,
  Information and Computation}, pages 212--217.

\bibitem[{Magistry and Sagot(2012)}]{magistry2012unsupervized}
Pierre Magistry and Beno{\^\i}t Sagot. 2012.
\newblock Unsupervized word segmentation: the case for mandarin chinese.
\newblock In \emph{Proceedings of the 50th Annual Meeting of the Association
  for Computational Linguistics: Short Papers-Volume 2}, pages 383--387.
  Association for Computational Linguistics.

\bibitem[{Magistry and Sagot(2013)}]{magistry2013can}
Pierre Magistry and Beno{\^\i}t Sagot. 2013.
\newblock Can mdl improve unsupervised chinese word segmentation?
\newblock In \emph{Sixth International Joint Conference on Natural Language
  Processing: Sighan workshop}, page~2.

\bibitem[{Mikolov et~al.(2013)Mikolov, Sutskever, Chen, Corrado, and
  Dean}]{mikolov2013distributed}
Tomas Mikolov, Ilya Sutskever, Kai Chen, Greg~S Corrado, and Jeff Dean. 2013.
\newblock Distributed representations of words and phrases and their
  compositionality.
\newblock In \emph{Advances in neural information processing systems}, pages
  3111--3119.

\bibitem[{Mochihashi et~al.(2009)Mochihashi, Yamada, and
  Ueda}]{mochihashi2009bayesian}
Daichi Mochihashi, Takeshi Yamada, and Naonori Ueda. 2009.
\newblock Bayesian unsupervised word segmentation with nested pitman-yor
  language modeling.
\newblock In \emph{Proceedings of the Joint Conference of the 47th Annual
  Meeting of the ACL and the 4th International Joint Conference on Natural
  Language Processing of the AFNLP: Volume 1-Volume 1}, pages 100--108.
  Association for Computational Linguistics.

\bibitem[{Pei et~al.(2014)Pei, Ge, and Chang}]{pei-ge-chang:2014:P14-1}
Wenzhe Pei, Tao Ge, and Baobao Chang. 2014.
\newblock Max-margin tensor neural network for chinese word segmentation.
\newblock In \emph{Proceedings of the 52nd Annual Meeting of the Association
  for Computational Linguistics (Volume 1: Long Papers)}, pages 293--303,
  Baltimore, Maryland. Association for Computational Linguistics.

\bibitem[{Sun(2010)}]{sun:2010:POSTERS}
Weiwei Sun. 2010.
\newblock Word-based and character-based word segmentation models: Comparison
  and combination.
\newblock In \emph{Coling 2010: Posters}, pages 1211--1219, Beijing, China.
  Coling 2010 Organizing Committee.

\bibitem[{Teh et~al.(2005)Teh, Jordan, Beal, and Blei}]{teh2005sharing}
Yee~W Teh, Michael~I Jordan, Matthew~J Beal, and David~M Blei. 2005.
\newblock Sharing clusters among related groups: Hierarchical dirichlet
  processes.
\newblock In \emph{Advances in neural information processing systems}, pages
  1385--1392.

\bibitem[{Wang et~al.(2017)Wang, Wang, Huang, Mohamed, Zhou, and
  Deng}]{wang2017sequence}
Chong Wang, Yining Wang, Po-Sen Huang, Abdelrahman Mohamed, Dengyong Zhou, and
  Li~Deng. 2017.
\newblock Sequence modeling via segmentations.
\newblock \emph{arXiv preprint arXiv:1702.07463}.

\bibitem[{Wang and Xu(2017)}]{Wang:2017}
Chunqi Wang and Bo~Xu. 2017.
\newblock {Convolutional Neural Network with Word Embeddings for Chinese Word
  Segmentation}.
\newblock In \emph{Proceedings of the 8th International Joint Conference on
  Natural Language Processing}.

\bibitem[{Wang et~al.(2011)Wang, Zhu, Tang, and Fan}]{wang2011new}
Hanshi Wang, Jian Zhu, Shiping Tang, and Xiaozhong Fan. 2011.
\newblock A new unsupervised approach to word segmentation.
\newblock \emph{Computational Linguistics}, 37(3):421--454.

\bibitem[{Yu et~al.(2016)Yu, Buys, and Blunsom}]{yu2016online}
Lei Yu, Jan Buys, and Phil Blunsom. 2016.
\newblock Online segment to segment neural transduction.
\newblock \emph{arXiv preprint arXiv:1609.08194}.

\bibitem[{Zhao and Kit(2008)}]{zhao2008empirical}
Hai Zhao and Chunyu Kit. 2008.
\newblock An empirical comparison of goodness measures for unsupervised chinese
  word segmentation with a unified framework.
\newblock In \emph{Proceedings of the Third International Joint Conference on
  Natural Language Processing: Volume-I}.

\end{thebibliography}
\bibliographystyle{acl_natbib_nourl}

\end{document}